\documentclass[letterpaper]{article} 
\usepackage{aaai24}  
\usepackage{times}  
\usepackage{helvet}  
\usepackage{courier}  
\usepackage[hyphens]{url}  
\usepackage{graphicx} 
\usepackage{natbib}  
\usepackage{caption} 
\usepackage{algorithm}
\usepackage{algorithmic}

\usepackage{newfloat}
\usepackage{listings}
\usepackage{amssymb}
\usepackage{amsmath}
\usepackage{makecell}
\usepackage{booktabs}
\usepackage{multirow}
\usepackage[usenames,dvipsnames]{color}

\usepackage{newfloat}
\usepackage{listings}
\DeclareCaptionStyle{ruled}{labelfont=normalfont,labelsep=colon,strut=off} 
\floatstyle{ruled}
\newfloat{listing}{tb}{lst}{}
\floatname{listing}{Listing}
\title{Depth-Guided Robust and Fast Point Cloud Fusion NeRF for Sparse Input Views}

\author {
    Shuai Guo\textsuperscript{\rm 1},
    Qiuwen Wang\textsuperscript{\rm 1},
    Yijie Gao\textsuperscript{\rm1},
    Rong Xie\textsuperscript{\rm 1},
    Li Song\textsuperscript{\rm 1,\rm 2\thanks{Corresponding author.}}
}
\affiliations {
    \textsuperscript{\rm 1}Institute of Image Communication and Network Engineering, Shanghai Jiao Tong University, Shanghai, China\\
    \textsuperscript{\rm 2}Cooperative Medianet Innovation Center, Shanghai Jiao Tong University, Shanghai, China\\
    \{shuaiguo, wangqiuwen, gaoyijie, xierong, song\_li\}@sjtu.edu.cn
}

\usepackage{bibentry}

\begin{document}

\maketitle

\begin{abstract}
Novel-view synthesis with sparse input views is important for real-world applications like AR/VR and autonomous driving.
Recent methods have integrated depth information into NeRFs for sparse input synthesis, leveraging depth prior for geometric and spatial understanding.
However, most existing works tend to overlook inaccuracies within depth maps and have low time efficiency.
To address these issues, we propose a depth-guided robust and fast point cloud fusion NeRF for sparse inputs.
We perceive radiance fields as an explicit voxel grid of features.
A point cloud is constructed for each input view, characterized within the voxel grid using matrices and vectors.
We accumulate the point cloud of each input view to construct the fused point cloud of the entire scene.
Each voxel determines its density and appearance by referring to the point cloud of the entire scene.
Through point cloud fusion and voxel grid fine-tuning, inaccuracies in depth values are refined or substituted by those from other views.
Moreover, our method can achieve faster reconstruction and greater compactness through effective vector-matrix decomposition.
Experimental results underline the superior performance and time efficiency of our approach compared to state-of-the-art baselines.
\end{abstract}

\section{Introduction}

Novel-view synthesis (NVS) serves as a fundamental objective within the realm of computer vision. 
The recent surge in NVS popularity is largely attributable to the success of Neural Radiance Fields (NeRFs)\cite{mildenhall2021nerf}.
However, NeRFs generally demand numerous images taken from a variety of views for efficient training.
In real-world applications such as AR/VR and autonomous driving, where input views are typically sparse~\cite{RegNeRF}, NeRF risks overfitting.
This may lead to inconsistencies in reconstructions or failure in generating any useful solution.

Various strategies have significantly enhanced the performance of NeRF for sparse inputs by 1) optimizing training data utilization~\cite{yu2021pixelnerf,RegNeRF,SPARF}, 2) incorporating prior information like depth and flow~\cite{DsNeRF,DDP}, or 3) exploring new constraints and regularizations~\cite{Infonerf,FreeNeRF}.
Among these solutions, depth prior has garnered substantial interest due to its ease of access, its capacity to offer object positions, and its aid in handling occlusions and geometry understanding.
Numerous methods have been proposed to integrate depth information into NeRFs for sparse input views.

\begin{figure}
\centering
\includegraphics[width = \linewidth]{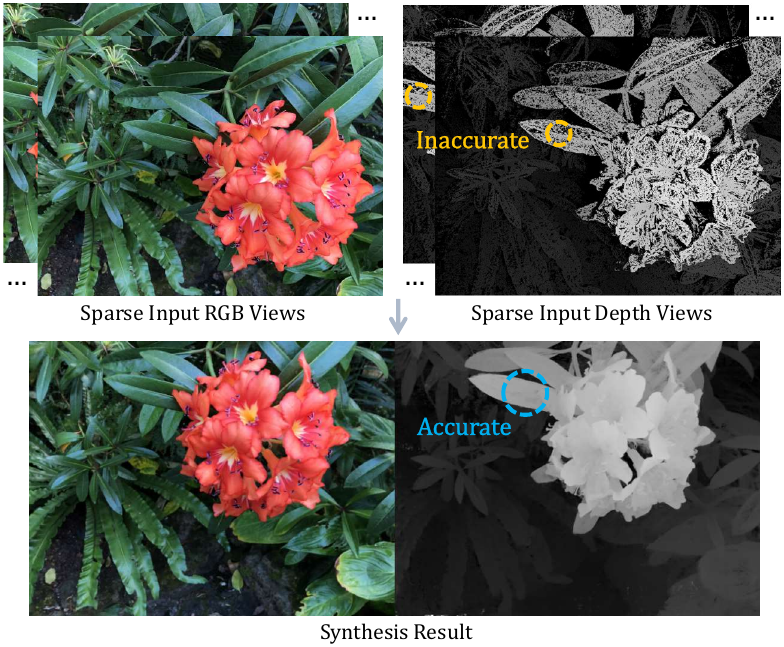}
\caption
{Depth-guided sparse input NeRF should overcome the effects of inaccurate depth values. This example illustrates a synthesis result of our method on the LLFF dataset.}
\label{inaccurateimg}
\end{figure}

Nevertheless, most existing depth-aware NeRFs for sparse input views disregard the holes, artifacts, and inaccurate values of depth maps.
For example, DSNeRF~\cite{DsNeRF} introduced depth supervision to leverage depth information, but it failed to account for inaccurate depth values.
Certain practices might introduce additional unreliability.
DDP-NeRF~\cite{DDP}, for instance, used a depth completion network to transform sparse depth into dense depth maps and uncertainty estimates, potentially leading to more inaccurate depth values.
The issues inherent to the network itself could subsequently impact sparse view synthesis effects.
SparseNeRF~\cite{SparseNeRF} employed the rough point cloud geometry provided by sparse RGB-D inputs to render more images and depict the approximate scene appearance.
However, the images rendered from sparse point clouds may be of low quality.

Moreover, few existing depth-aware NeRFs have used depth information to create faster NeRFs, resulting in overall low time efficiency.
For instance, DSNeRF~\cite{DsNeRF}, despite its claims of improved speed, required several hours of training.
On the other hand, Mip-NeRF~\cite{dey2022mip}, which employed depth supervision and depth-assisted local sampling, managed to train 3-5 times faster, but its training duration still approached an hour.

To address the challenges above, we introduce a depth-guided robust and fast point cloud fusion NeRF tailored for sparse input views.
This is the first integration of point cloud fusion with NeRF volumetric rendering.
In particular, inspired by TensoRF~\cite{chen2022tensorf}, we perceive radiance fields as an explicit voxel grid of features, delineated by a series of vectors and matrices that articulate scene appearance and geometry along their respective axes.
The feature grid can be naturally seen as a 4D tensor, where three of its modes correspond to the XYZ axes of the grid, and the fourth mode represents the feature channel dimension.
Utilizing sparse input RGB-D images and camera parameters, we map the 2D pixels of each input view to 3D space to generate a point cloud for each view.
Subsequently, we convert depth values into densities, and encode both the depth and color information into the voxel grid utilizing two distinct sets of matrices and vectors.
Volume density and view-dependent color can be decoded from the features, facilitating volumetric radiance field rendering.
We aggregate the point cloud from each input view to assemble the fused point cloud of the entire scene.
Each voxel determines its density and appearance within the scene by referencing this fused point cloud.

\begin{figure}
\centering
\includegraphics[width = \linewidth]{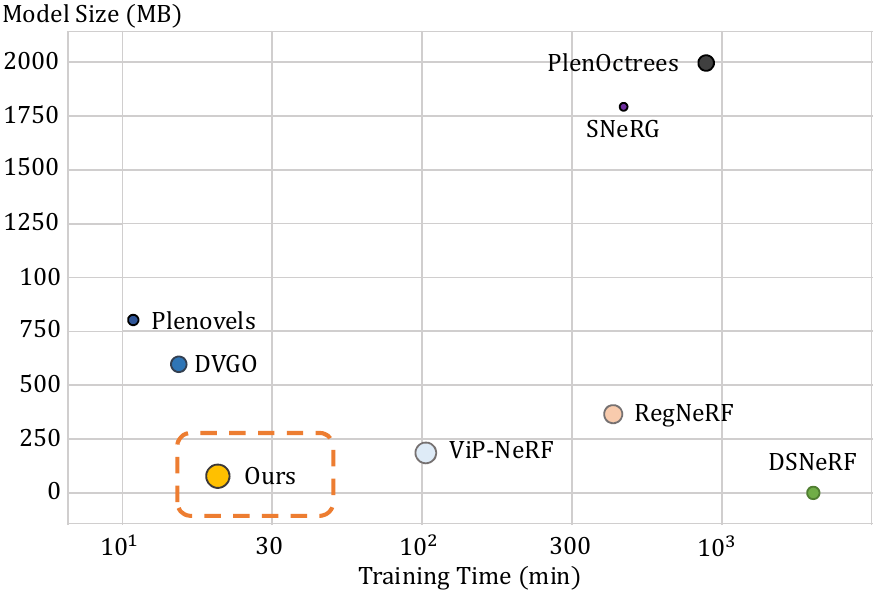}
\caption
{We compare our method with previous methods in terms of rendering quality (PSNR) and model size. Point sizes correspond to PNSRs. With effective vector-matrix decomposition and point cloud presentation, our work delivers superior rendering quality, faster reconstruction, and greater compactness.}
\label{fast}
\end{figure}

Since the planes and vectors are iteratively refined during the training process, and the point cloud of the entire scene is composed of all input views, inaccurate depth values are refined or replaced by depth values from other views.
Figure~\ref{inaccurateimg} illustrates a synthesis example of our method.
Additionally, as the vector-matrix decomposition technique effectively minimizes the number of components needed for the same expression capacity, our method can achieve faster reconstruction and greater compactness, as shown in Figure~\ref{fast}.
This paper primarily contributes the following:
\begin{itemize}
\item We introduce the first depth-guided robust and fast point cloud fusion NeRF for sparse view input, minimizing the impact of inaccurate depth values.
\item To our knowledge, this is the first NeRF framework that is integrated with point cloud fusion, offering a novel NeRF scene representation.
\item Our method boosts time efficiency, and delivers superior results compared to state-of-the-art methods.
\end{itemize}

\section{Related Work}
In this section, we provide a comprehensive review of the relevant literature in the areas of novel-view synthesis and sparse input NeRF.

\subsection{Novel-View Synthesis}
The body of work about novel-view synthesis can generally be divided into two main categories: explicit representation based synthesis and implicit representation based synthesis.

\subsubsection{Explicit Representations}
Explicit representation methods commonly employ point clouds~\cite{ran2022surface,huang2023boosting}, voxels~\cite{sitzmann2019deepvoxels,song2022jpv}, meshes~\cite{feng2019meshnet,yang2023dmis}, or MPI~\cite{zhou2018stereo,kundu2020kinematic} to represent 3D geometry and appearance.
Despite their computational efficiency, these techniques often pose optimization challenges due to their discontinuous nature.

\subsubsection{Implicit Representations}
Implicit methods directly model the appearance of a 3D scene, thus eliminating the need for explicit geometric representation.
A prime example of this approach is NeRF~\cite{mildenhall2021nerf}.
NeRF~\cite{mildenhall2021nerf} assigns a color and opacity to a given 3D location and 2D viewing direction, which correspond to the light emitted from that specific location in that particular direction.
Owing to its inherent simplicity and superior rendering quality, NeRF has been widely adopted in recent studies for numerous extensions, including but not limited to, video synthesis~\cite{li2022streaming,li2022neural}, relighting~\cite{yu2022anisotropic,rudnev2022nerf}, and scene editing~\cite{yuan2022nerf,kobayashi2022decomposing}.

\begin{figure*}[t]
\centering
\includegraphics[width = \linewidth]{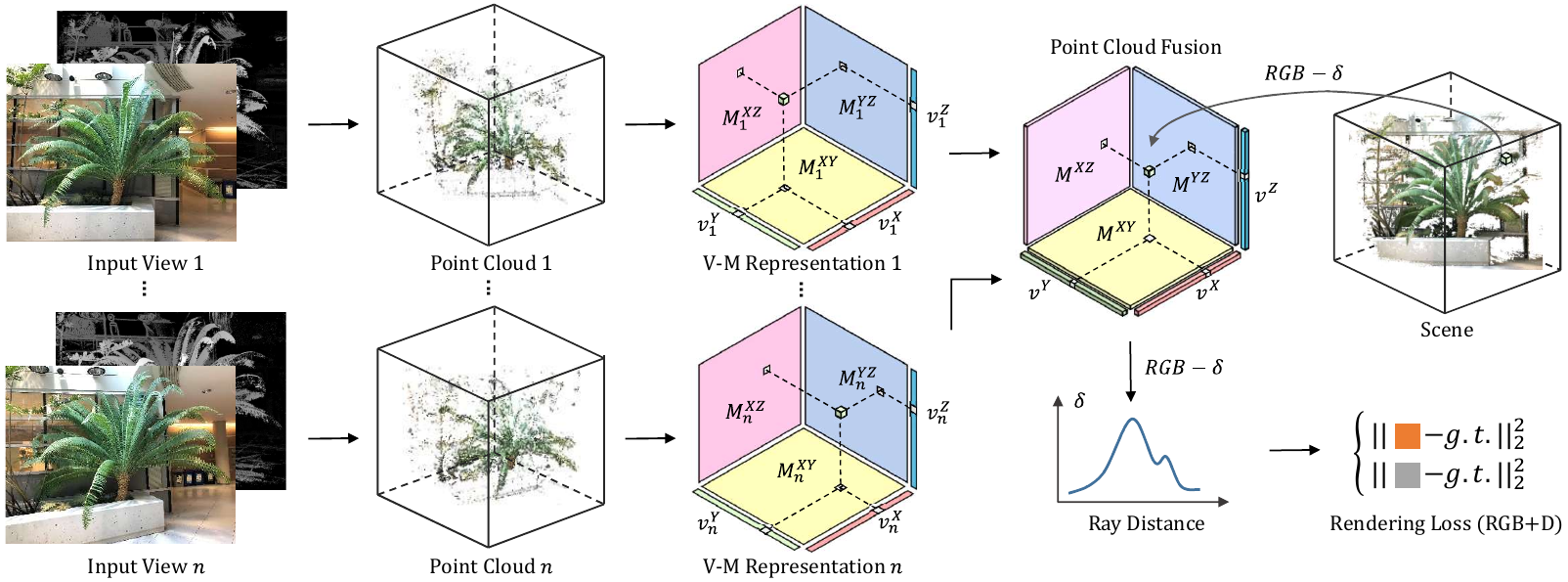}
\caption
{Overview of our method. We perceive radiance fields as an explicit voxel grid of features. With RGB-D images and camera parameters of $n$ sparse input views, we first map pixel points into 3D space to construct a point cloud for each view, represented by vectors and matrices. Then we accumulate the point cloud of each input view to construct the fused point cloud of the entire scene. For each shading location $\mathrm{x}_w=(x,y,z)$, we use sampled values from the vectors and matrices to compute the corresponding values of the tensor component. The appearance values are sent to a decoding MLP $S$ for color regression. The loss function is composed of RGB loss and depth loss.}
\label{overview}
\end{figure*}

\subsection{Sparse Input NeRF}

A number of works have been proposed to address the data-hungry problem of NeRF by exploiting training data, incorporating prior information like depth and flow, or introducing new constraints and regularizations.

\subsubsection{Exploiting Training Data}
PixelNeRF~\cite{yu2021pixelnerf} enhances scene comprehension by conditioning a NeRF on image inputs using a fully convolutional approach.
RegNeRF~\cite{RegNeRF} applies regularization to the geometry and appearance of rendered patches from unseen views, helping to rectify inaccurately optimized scene geometry and divergent behavior at the optimization outset.
SPARF~\cite{SPARF} employs pixel matches between input views and depth consistency to generate realistic novel-view renderings with sparse inputs. 

\subsubsection{Leveraging Prior Information}
DSNeRF~\cite{DsNeRF}, for example, uses the sparse depth information created by COLMAP~\cite{COLMAP} as explicit supervision for sparse view synthesis.
SparseNeRF~\cite{SparseNeRF}, meanwhile, relies on depth ranking prior and spatial continuity distillation on NeRFs, enabling the synthesis of novel views with sparse view inputs.
DDP-NeRF~\cite{DDP} synthesizes novel views of entire rooms from significantly fewer images by employing dense depth priors to constrain the NeRF optimization, thereby enabling data-efficient novel-view synthesis on challenging indoor scenes.

\subsubsection{Introducing New Regularizations}
InfoNeRF~\cite{Infonerf} enhances the compactness of reconstructed scenes along individual rays while ensuring consistency across neighboring rays, an approach particularly beneficial for few-shot novel-view synthesis. 
Drawing on the importance of frequency, FreeNeRF~\cite{FreeNeRF} regulates the frequency range of NeRF's inputs. It also imposes penalties on the density fields near the camera.
Both methods demonstrate innovative approaches to manipulating input parameters to optimize output with sparse views.

\section{Method}
We first revisit the feature grids and radiance field, followed by an analysis of factorizing radiance fields and point cloud representation.
Subsequently, we illustrate the process of continuous field representation, point cloud fusion, and rendering.
The presentation concludes with a discussion on the optimization and loss function.
Figure~\ref{overview} is an overview of our method.

\subsection{Feature Grids and Radiance Field Revisited}

We construct a model of a radiance field which establishes a relationship between a 3D location $x$ and a viewing direction $d$, with its volume density $\sigma$, and a view-dependent color $c$.
Taking inspiration from TensoRF~\cite{chen2022tensorf}, we employ a standard 3D grid $\mathcal{G}\in \mathbb{R}^{I\times J\times K}$.
Each voxel within this grid is equipped with multi-channel features, thus allowing us to simulate this function.
Here, $I$, $J$, and $K$ represent the resolutions of the feature grid along the X, Y, and Z axes, respectively.
We segregate these feature channels into two distinct grids, one for geometry, represented by $\mathcal{G}_\sigma \in \mathbb{R}^{I\times J\times K}$, and another for appearance, represented by $\mathcal{G}_c\in \mathbb{R}^{I\times J\times K\times P}$.
In this context, $P$ indicates the number of appearance feature channels.
These individual grids are designed to separately model the volume density $\sigma$, and the view-dependent color $c$.

Our model accommodates a range of appearance features within the appearance grid $\mathcal{G}_c$, which depend on a predefined function $S$.
This function transforms an appearance feature vector in combination with a viewing direction $d$, into a color $c$.
Here $S$ is a small MLP where the appearance grid $\mathcal{G}_c$ comprises neural features and spherical harmonics (SH) coefficients, respectively.
On the other hand, we introduce a single-channel grid $\mathcal{G}_\sigma$, where the values directly represent volume density, thus eliminating the need for an extra conversion function.
This results in a continuous grid-based radiance field which can be expressed by the equation:
\begin{equation}
\sigma, c = \mathcal{G}_\sigma(x),S(\mathcal{G}_c(x),d).
\label{eq1}
\end{equation}
In this equation, $\mathcal{G}_\sigma(x)$ and $\mathcal{G}_c(x)$ represent the features from the two grids at the location $x$, interpolated trilinearly.

\subsection{Factorizing Radiance Fields}
We model the geometry grid $\mathcal{G}_\sigma$ and the appearance grid $\mathcal{G}_c$ as factorized tensors.
Utilizing the Vector-Matrix (VM) decomposition, we factorize the 3D geometry tensor $\mathcal{G}_\sigma$ as:
\begin{align}
\begin{split}
\mathcal{G}_\sigma = \sum_{r=1}^{n} v_{\sigma,r}^X \circ M_{\sigma,r}^{YZ}& + v_{\sigma,r}^Y \circ M_{\sigma,r}^{XZ} + v_{\sigma,r}^Z \circ M_{\sigma,r}^{XY} \\
= \sum_{r=1}^{n}& \sum_{m\in XYZ}\mathcal{A}_{\sigma, r}^m,
\end{split}
\end{align}
where $v_{\sigma,r}^X \in \mathbb{R}^I$, $v_{\sigma,r}^Y \in \mathbb{R}^J$, $v_{\sigma,r}^Z \in \mathbb{R}^K$, $M_{\sigma,r}^{XY}\in \mathbb{R}^{I\times J}$, $M_{\sigma,r}^{YZ}\in \mathbb{R}^{J\times K}$, and $M_{\sigma,r}^{XZ}\in \mathbb{R}^{I\times K}$.

Our approach differs from TensoRF in that the number of components of the 3D geometry tensor in our method is consistently fixed as the number of input views, denoted as $n$.
Each component corresponds to an input view and is used to represent the volume density of the point cloud, constructed by this view.
Similarly, the appearance tensor $\mathcal{G}_c$ is modeled using comparable vector-matrix spatial factors and additional feature basis vectors $b_r$, which express a multi-channel voxel feature grid:
\begin{align}
\begin{split}
\mathcal{G}_c = \sum_{r=1}^{n}\mathcal{A}_{c, r}^X \circ b_{3r-2}& + \mathcal{A}_{c, r}^Y \circ b_{3r-1} + \mathcal{A}_{c, r}^Z \circ b_{3r} ,\\
\mathcal{A}_{c, r}^X =& v_{c,r}^X \circ M_{c,r}^{YZ},\\ 
\mathcal{A}_{c, r}^Y =& v_{c,r}^Y \circ M_{c,r}^{XZ},\\
\mathcal{A}_{c, r}^Z =& v_{c,r}^Z \circ M_{c,r}^{XY}.
\end{split}
\end{align}
In this context, $v_{c,r}^X \in \mathbb{R}^I$, $v_{c,r}^Y \in \mathbb{R}^J$, $v_{c,r}^Z \in \mathbb{R}^K$, $M_{c,r}^{XY}\in \mathbb{R}^{X\times Y}$, $M_{c,r}^{YZ}\in \mathbb{R}^{J\times K}$, and $M_{c,r}^{XZ}\in \mathbb{R}^{I\times K}$.

Contrasting with TensoRF, our method also fixes the number of components of the appearance tensor as the number of input views $n$. 
Here, each component corresponds to an input view and represents the appearance of the point cloud constructed from this view.
We maintain $3n$ vectors $b_r$ to match the total number of components.
By stacking all $b_r$, we create a $P\times 3n$ matrix $B$, which serves as a global appearance dictionary, abstracting the appearance commonalities across the entire scene.
A density value $\mathcal{G}_{\sigma,ijk}$ of a single voxel at indices $ijk$ can be calculated by the provided equation:
\begin{equation}
\mathcal{G}_{\sigma,ijk} = \sum_{r=1}^{n} \sum_{m\in XYZ}\mathcal{A}_{\sigma, r\atop ijk}^m.
\end{equation}
In parallel, the appearance grid $\mathcal{G}_{c,ijk}$, corresponding to $\mathcal{G}_c$ at fixed XYZ indices $ijk$, can also be calculated by the given method:
\begin{align}
\begin{split}
\mathcal{G}_{c,ijk} = \sum_{r=1}^{n}\mathcal{A}_{c, r\atop ijk}^X &\circ b_{3r-2} + \mathcal{A}_{c, r\atop ijk}^Y \circ b_{3r-1} \\
+ \mathcal{A}_{c, r\atop ijk}^Z \circ b_{3r} &=  B\oplus([\mathcal{A}_{\sigma, r\atop ijk}^m]_{m,r}).
\end{split}
\end{align}

\subsection{Point Cloud Representation}

We construct a point cloud for each input view, which is represented within the voxel grid using the corresponding feature vectors and matrices.
Initially, we map depth values to the normalized device coordinate (NDC) space to ensure that all visible locations are normalized and represented within a predetermined cubic space.
Subsequently, we map all pixels of the input view to the 3D space to generate the point cloud.
For a 2D pixel location, $\mathrm{x}_p$, in the $r-$th input image, we map it to a 3D world location $\mathrm{x}_w=(x,y,z)$, using the equation:
\begin{equation}
\mathrm{x}_w = R^{-1}(K^{-1}\mathrm{x}_pD-t),
\end{equation}
where $K$, $R$, and $t$ denote the intrinsic parameters, rotation matrix, and translation matrix of the input view corresponding to $\mathrm{x}_p$.

Next, we represent the point cloud by constructing its geometry grid $\mathcal{G}_{\sigma}$ and appearance grid $\mathcal{G}_{c}$.
As the 3D space is represented by a standard 3D grid $\mathcal{G}\in \mathbb{R}^{I\times J\times K}$, the world location $\mathrm{x}_w=(x,y,z)$ is projected onto the matrix $M_{\sigma,r}^{XY},M_{\sigma,r}^{YZ},M_{\sigma,r}^{XZ}$ and vectors $v_{\sigma,r}^X,v_{\sigma,r}^Y,v_{\sigma,r}^Z$.
Elements of the matrices and vectors that are projections of the point cloud are assigned a value of 1, while all other elements are assigned a value of 0.
This assignment method effectively indicates the presence of points in the geometry grid.

For the appearance grid, elements of the matrix $M_{c,r}^{XY}$, $M_{c,r}^{YZ}$, $M_{c,r}^{XZ}$ are assigned the average of R, G, B color values of the points projected onto this element.
Meanwhile, vectors $v_{c,r}^X$, $v_{c,r}^Y$, and $v_{c,r}^Z$ are assigned random values.
This provides a rough representation of the point clouds.
After the training step, the representation of point clouds will be refined to overcome inaccurate depth values and will be fused together to characterize the entire scene.

\subsection{Continuous Field Representation}

We employ trilinear interpolation to represent a continuous field.
For instance, consider a component tensor represented as $\mathcal{A}_r^X = v_r^X\circ M_r^{YZ}$.
Each tensor element within this can be described as $\mathcal{A}_{r,ijk}^X = v_{r,i}^X M_{r,jk}^{YZ}$.
The interpolated values can then be calculated using:
\begin{equation}
\mathcal{A}_r^X (\mathrm{x}) = v_r^X(x) M_r^{YZ}(y,z).
\end{equation}
In the equation above, $\mathcal{A}_r^X(\mathrm{x})$ signifies the trilinearly interpolated value at the 3D location $\mathrm{x}=(x,y,z)$ of $\mathcal{A}_r$.
The term $v_r^X(x)$ represents the linear interpolation at position $x$ along the X-axis.
Meanwhile, $M_r^{YZ}(y,z)$ denotes the bilinear interpolation at $(y,z)$ of $M_r^{YZ}$ in the YZ plane. 
Similarly, the following relations hold:

\begin{align}
\begin{split}
\mathcal{A}_r^Y (\mathrm{x}) = v_r^Y(y) M_r^{XZ}(x,z),\\
\mathcal{A}_r^Z (\mathrm{x}) = v_r^Z(z) M_r^{XY}(x,y).
\end{split}
\end{align}

By trilinearly interpolating both grids and merging the point cloud, we obtain:

\begin{equation}
\mathcal{G}_{\sigma}(\mathrm{x}) = \sum_r\sum_m \mathcal{A}_{\sigma,r}^m(\mathrm{x}),
\label{eq13}
\end{equation}
\begin{equation}
\mathcal{G}_{c}(\mathrm{x}) = B\oplus([\mathcal{A}_{c,r}^m(\mathrm{x})]_{m,r}).
\label{eq14}
\end{equation}

\subsection{Point Cloud Fusion and Rendering}
Integrating equations (\ref{eq1}), (\ref{eq13}) and (\ref{eq14}), the factorized tensorial radiance field for the fused point cloud in our model is articulated as:
\begin{equation}
\sigma,c = \sum_r\sum_m \mathcal{A}_{\sigma,r}^m(x), S(B\oplus([\mathcal{A}_{c,r}^m(\mathrm{x})]_{m,r}),d).
\end{equation}

To render images, we march along a ray, and $Q$ shading points are sampled along each ray.
The color of the pixel is then determined using:
\begin{align}
\begin{split}
C = \sum_{q=1}^Q\tau_q(1-&\exp(-\sigma_q\Delta_q))c_q,\\
\tau_q=\exp(&-\sum_{p=1}^{q-1}\sigma_p\Delta_p).
\end{split}
\end{align}
Here, $\sigma_q$ and $c_q$ denote the density and color respectively, determined by our model at their specific sampled locations $\mathrm{x}_q$.
Meanwhile, $\Delta_q$ is defined as the step size of the ray and $\tau_q$ stands for transmittance.

To further improve quality and avoid local minima, we apply coarse-to-fine reconstruction.
Similar to TensoRF~\cite{chen2022tensorf}, our coarse-to-fine reconstruction is simply achieved by linearly and bilinearly upsampling our XYZ-mode vector and matrix factors.

\subsection{Optimization}

The network parameters are optimized using a collection of RGB-D frames, each containing color, depth, and camera pose data.
Our loss function consists of two primary components.
The first component is an RGB loss function.
This component involves an L2 rendering loss combined with additional regularization terms to optimize our tensor factors for radiance field reconstruction. It can be represented as:

\begin{equation}
\mathcal{L}_{RGB} = \|C-\widetilde{C}\|_2^2 + \omega_{reg} \mathcal{L}_{reg}.
\end{equation}
Here, $\widetilde{C}$ denotes the ground truth color, $C$ represents the predicted color, $\mathcal{L}_{reg}$ is an L1 regularization term, and $\omega_{reg}$ is the weight assigned to this regularization.
To promote sparsity in our tensor factors' parameters, we use the standard L1 regularization.
This technique has proven effective in enhancing the extrapolation of views and eliminating anomalies like floaters or outliers in the final renderings.

\begin{figure}[t]
\centering
\includegraphics[width = \linewidth]{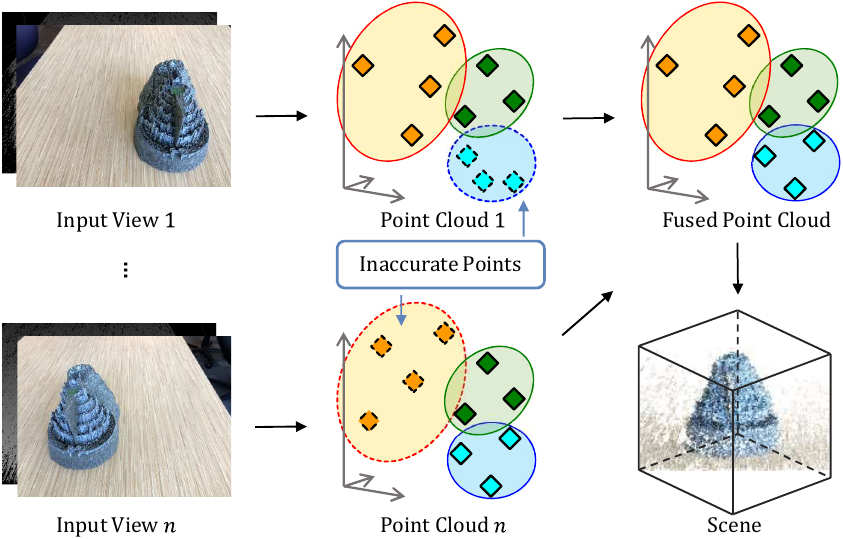}
\caption
{Inaccurate depth values can result in inaccurate 3D points. Through point cloud fusion and radiance field optimization, these inaccurate 3D points are substituted with accurate ones from other views. The squares represented by the dotted edges indicate inaccurate 3D points.}
\label{fusion}
\end{figure}

The second component is a depth loss function, in which we employ an L2 rendering loss and can be expressed as:
\begin{equation}
\mathcal{L}_{depth} = \|D-\widetilde{D}\|_2^2.
\end{equation}
In this equation, $\widetilde{D}$ is the ground truth depth, while $D$ stands for the predicted depth.
Contrary to the approach in DSNeRF~\cite{DsNeRF}, we normalize the ground truth depth in the NDC space.
This normalization ensures that all visible locations are encapsulated within a predefined cubic space.

In summary, the composite loss function for our model, denoted by $\mathcal{L}$, is articulated as:
\begin{equation}
\mathcal{L} = \mathcal{L}_{RGB} + \lambda_{depth} \mathcal{L}_{depth},
\label{eqloss}
\end{equation}
where $\lambda_{depth}$ serves as a hyperparameter to strike a balance between color and depth supervision.
Figure~\ref{fusion} illustrates how the inaccuracies are addressed through point cloud fusion and radiance field optimization.

\begin{table*}[h]
\centering
\begin{tabular}{cp{1.3cm}<{\centering}p{1.3cm}<{\centering}p{1.3cm}<{\centering}p{1.3cm}<{\centering}p{1.3cm}<{\centering}p{1.3cm}<{\centering}p{1.3cm}<{\centering}p{1.3cm}<{\centering}p{1.3cm}<{\centering}}
\hline
\multirow{2}*{Method} & \multicolumn{3}{c}{2 views} & \multicolumn{3}{c}{3 views} & \multicolumn{3}{c}{4 views} \\
~ & PSNR$\uparrow$ & SSIM$\uparrow$ & LPIPS$\downarrow$ & PSNR$\uparrow$ & SSIM$\uparrow$ & LPIPS$\downarrow$ & PSNR$\uparrow$ & SSIM$\uparrow$ & LPIPS$\downarrow$ \\
\hline
InfoNeRF & 9.23 & 0.2095 & 0.7761 & 8.52 & 0.1859 & 0.7679 & 9.25 & 0.2188 & 0.7701 \\
DietNeRF & 11.89 & 0.3209 & 0.7265 & 11.77 & 0.3297 & 0.7254 & 11.84 & 0.3404 & 0.7396 \\
RegNeRF & 16.90 & 0.4872 & 0.4402 & 18.62 & 0.5600 & 0.3800 & 19.83 & 0.6056 & 0.3446 \\
DSNeRF & 17.06 & 0.5068 & 0.4548 & 19.02 & 0.5686 & 0.4077 & 20.11 & 0.6016 & 0.3825 \\
DDP-NeRF & 17.21 & 0.5377 & 0.4223 & 17.90 & 0.5610 & 0.4178 & 19.19 & 0.5999 & 0.3821 \\
ViP-NeRF & 16.76 & 0.5222 & 0.4017 & 18.92 & 0.5837 & 0.3750 & 19.57 & \textbf{0.6085} & 0.3593 \\
Our Method & \textbf{17.83} & \textbf{0.5512} & \textbf{0.3832} & \textbf{19.30} & \textbf{0.6027} & \textbf{0.3682} & \textbf{20.86} & 0.5967 & \textbf{0.3247} \\
\hline
\end{tabular}
\caption{Quantitative Comparisons on The LLFF Dataset.}
\label{llfftable}
\end{table*}

\begin{figure*}[t]
\centering
\includegraphics[width = \linewidth]{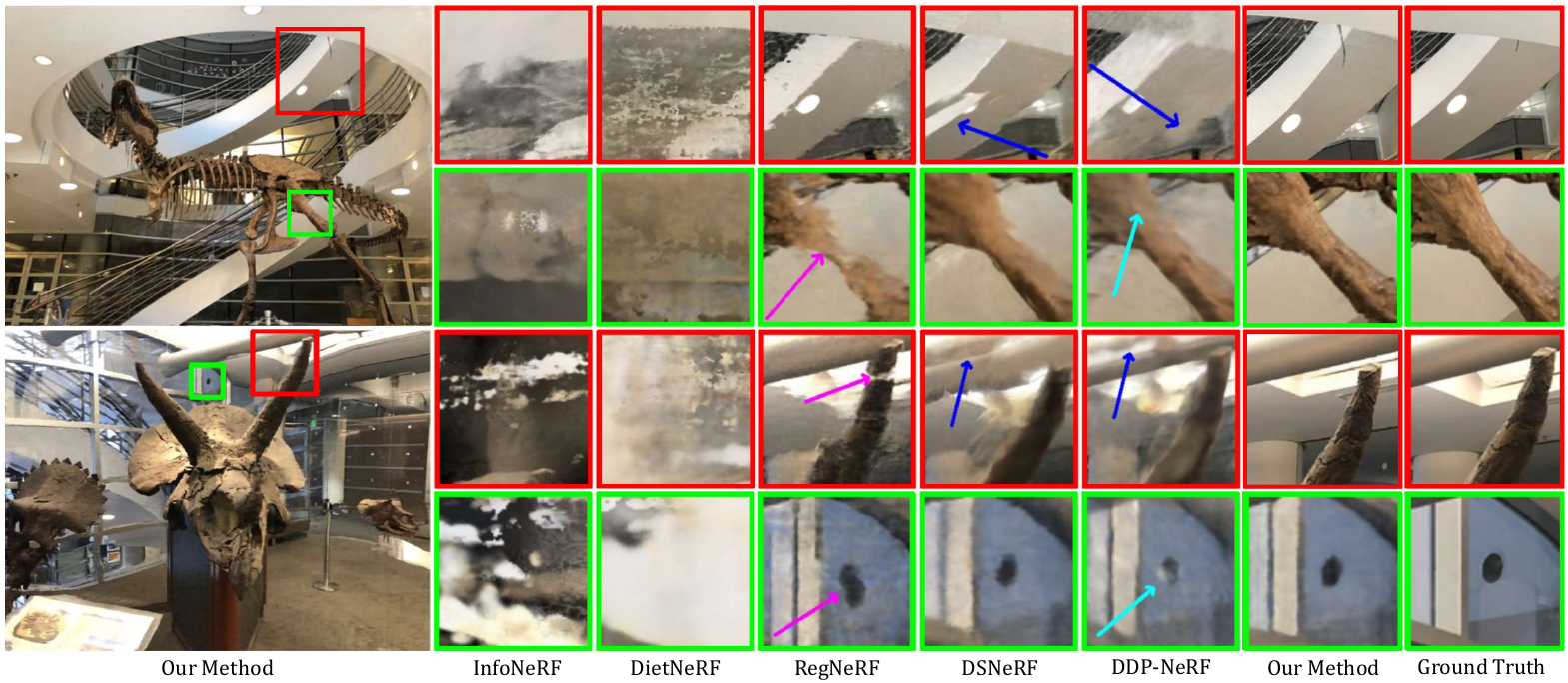}
\caption
{Qualitative comparisons on the LLFF dataset with two input views. Notably, the predictions from DSNeRF and DDP-NeRF exhibit noticeable floater artifacts. RegNeRF struggles to capture the finer details in bone structures. In contrast, our method significantly reduces these imperfections. In the second and fourth examples, we highlight the color changes predicted by DDP-NeRF. Our model's predictions are free from the aforementioned artifacts.}
\label{llffimage}
\end{figure*}

\section{Experiments}


\subsection{Datasets}
Our results are derived from the real-world multi-view datasets LLFF~\cite{mildenhall2019local} and DTU~\cite{jensen2014large}.
The LLFF dataset includes 8 forward-facing scenes, each having a varied count of frames, all presented at a spatial resolution of 4032×3024.
The DTU dataset features various objects captured from multiple perspectives in a controlled indoor setting. 

For every scene, we selected subsets containing 2, 3, or 4 training views for our evaluation.
Depth maps were generated using COLMAP~\cite{COLMAP}, adhering to the dataset-provided camera parameters.

\subsection{Baselines}
We compare our method against several state-of-the-art models, including InfoNeRF~\cite{Infonerf}, DietNeRF~\cite{jain2021putting}, RegNeRF~\cite{RegNeRF}, DSNeRF~\cite{DsNeRF}, DDP-NeRF~\cite{DDP}, and ViP-NeRF~\cite{somraj2023vip}.
It is noteworthy that all results of these methods are obtained through publicly accessible codes or papers.

\subsection{Implementation Details}
Our implementation was carried out in PyTorch~\cite{paszke2019pytorch}, excluding any customized CUDA kernels.
This model was optimized over $T$ iterations, with a batch size of 4096 pixel rays, executed on a single NVIDIA RTX 4090 GPU (24GB).
We introduced a feature decoding MLP and set $P=27$.
To facilitate a stepwise transition from coarse-to-fine reconstruction, we initiated with a grid of $N_0^3$, where $N_0=128$.
This grid was upsampled at intervals of 2000, 3000, 4000, 5500, and 7000 steps, with voxel counts transitioning linearly in logarithmic space from $N_0^3$ to $N^3$.

\begin{table*}[h]
\centering
\begin{tabular}{cp{1.3cm}<{\centering}p{1.3cm}<{\centering}p{1.3cm}<{\centering}p{1.3cm}<{\centering}p{1.3cm}<{\centering}p{1.3cm}<{\centering}p{1.3cm}<{\centering}p{1.3cm}<{\centering}p{1.3cm}<{\centering}}
\hline
\multirow{2}*{Dataset} & \multicolumn{3}{c}{0\% Noise} & \multicolumn{3}{c}{5\% Noise} & \multicolumn{3}{c}{10\% Noise} \\
~ & PSNR$\uparrow$ & SSIM$\uparrow$ & LPIPS$\downarrow$ & PSNR$\uparrow$ & SSIM$\uparrow$ & LPIPS$\downarrow$ & PSNR$\uparrow$ & SSIM$\uparrow$ & LPIPS$\downarrow$ \\
\hline
LLFF & 17.83 & 0.5512 & 0.3832 & 15.98 & 0.4058 & 0.4597 & 13.26 & 0.2615 & 0.6433 \\
DTU & 19.34 & 0.7431 & 0.2576 & 18.17 & 0.5976 & 0.3814 & 16.63 & 0.4144 & 0.5561 \\
\hline
\end{tabular}
\caption{Influence of depth quality on our method. In this table, we add 5\% and 10\% white noise to depth maps respectively to observe the performance of our method. The original depth maps are obtained with COLMAP.}
\label{depthquality}
\end{table*}

\subsection{Comparisons}

\subsubsection{Comparisons on The LLFF Dataset}
The quantitative and qualitative comparisons with competing models are respectively showcased in the referenced Table~\ref{llfftable} and Figure~\ref{llffimage}.
Here, we note that predictions from other models frequently exhibit blurriness, especially for views substantially distanced from the input.
In contrast, our method performs well in geometry predictions, generating more realistic novel views.
Our method consistently delivers sharp outputs accompanied by precise scene geometry across all tested scenarios.
A standout feature of our approach is its ability to accurately represent and recover intricate details.
Additionally, our model is good at eliminating visual noise, ensuring clearer visuals around objects in comparison to the baseline models.

\subsubsection{Comparisons on The DTU Dataset}
\begin{figure}
\centering
\includegraphics[width = \linewidth]{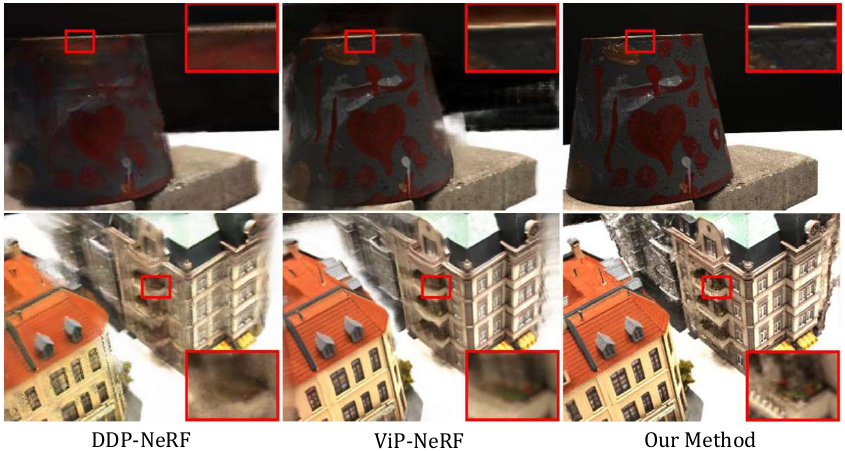}
\caption
{Qualitative comparisons on the DTU dataset with two input views. Predictions of DDP-NeRF and ViP-NeRF display pronounced floating cloud artifacts. Our method yields more lifelike and convincing novel views.}
\label{dtuimage}
\end{figure}

The qualitative comparisons between the competing models can be found in Figure~\ref{dtuimage}.
Our approach consistently surpasses other models, especially in the perceptual metric domain.
Additionally, these models often present inconsistent appearances for novel views, particularly when the camera perspectives deviate significantly from the input views.
We can see that our model aligns more closely with the ground truth and avoids many of the artifacts evident in the predictions of other models.

\subsection{Ablation Study}

\subsubsection{Ablation of Point Cloud Construction and Fusion}
\begin{figure}
\centering
\includegraphics[width = \linewidth]{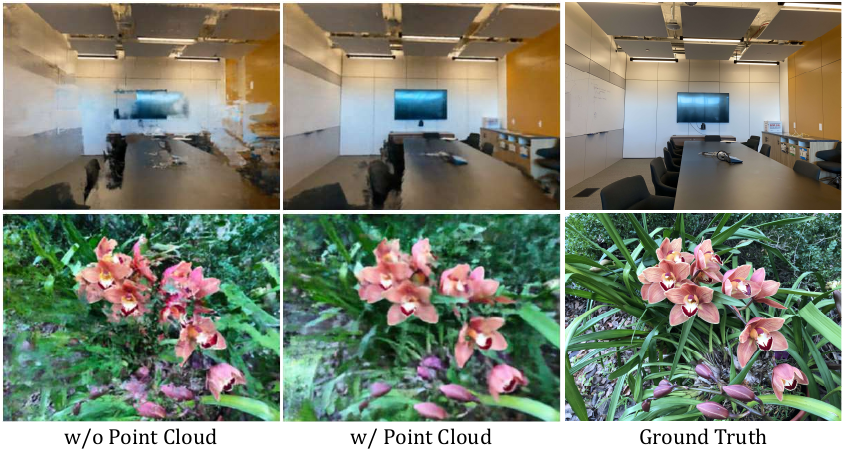}
\caption
{Influence of point cloud construction of our method. We show some qualitative examples on the LLFF dataset with two input views. Our method yields more lifelike and convincing novel views when point cloud construction and fusion are introduced.}
\label{pointcloudornot}
\end{figure}

In Figure~\ref{pointcloudornot}, we highlight the significance of our point cloud construction and fusion, which effectively addresses the challenges posed by inaccurate depth values.
When the point cloud construction is not applied, all matrices and vectors default to random values.
The adoption of our point cloud construction and fusion approach yields superior quantitative and qualitative outcomes.

\subsubsection{Depth Quality}
To delve deeper into the effects of varied depth quality on our method, we present synthesis outcomes as depth quality fluctuates in Table~\ref{depthquality}.
For reference, the highest quality depth maps are sourced from COLMAP~\cite{COLMAP}.
We intentionally degrade the quality of these depth maps by converting 5\% and 10\% of the best depth values into white noise.
Even with diminishing depth map quality, our method maintains commendable performance.

\subsection{Discussion}

With the depth supervision and our streamlined tensorial radiance field structure, our method boasts better performance and faster reconstruction.

To reduce the influence of inaccurate depth values, we use depth information to optimize the neural radiance field.
We project points from each input view into 3D space, creating a unique point cloud for every view, represented by matrices and vectors.
Although the initial representation of the point clouds is rudimentary and imprecise, the fusion and training processes refine inaccuracies in depth values or replace them with values from alternate views.

As shown in Figure~\ref{fast}, our method has less model size and less reconstruction time, as we effectively present the point cloud constructed for each input view with a few vectors and matrices.
For example, for a $300\times 300\times 300$ feature grid with $P=27$ channels (plus one density channel), the total number of parameters in a dense grid is 756 M, while the number of parameters used for our method is only about 0.36 M (four views input).
We can achieve a compression rate of about $0.05\%$.

\section{Conclusion}

In this paper, we introduce the pioneering depth-guided robust and fast point cloud fusion NeRF tailored for sparse view input.
We observed that existing depth-guided NeRFs for sparse input views tend to neglect inaccuracies in depth maps and often suffer from low time efficiency.
To the best of our knowledge, this represents the first integration of point cloud fusion into the NeRF framework.
Our method leverage depth information to construct a superior radiance field while reducing the influence of inaccurate depth values.
It also enables faster reconstruction and greater compactness via efficient vector-matrix decomposition.

\subsection{Limitations and FutureWork}
We believe that depth-guided radiance fields based on matrix and vector representations hold significant promise in time efficiency enhancement.
Moving forward, we will aim to further leverage depth information and tensorial structures to improve the performance and efficiency of NeRF.

\section{Acknowledgments}
This work was supported by the Fundamental Research Funds for the Central Universities, National Key R\&D Project of China (2019YFB1802701), MoE-China Mobile Research Fund Project (MCM20180702), STCSM under Grant 22DZ2229005, 111 project BP0719010.

\bigskip

\bibliography{aaai24}

\end{document}